\documentclass[conf]{new-aiaa}
\usepackage[utf8]{inputenc}

\usepackage{graphicx}
\usepackage{amsmath}
\usepackage[version=4]{mhchem}
\usepackage{siunitx}
\usepackage{longtable,tabularx}
\usepackage{subcaption}

\usepackage{mwe}
\usepackage{listings}
\usepackage{enumitem}
\usepackage{ulem}

\title{Visual Language Models as Operator Agents in the Space Domain}

\author{
    Alejandro Carrasco\footnote{B.S. in Computer Science and Engineering, \href{mailto:acarra@mit.edu}{acarra@mit.edu}}, 
    Marco Nedungadi\footnote{S.M. Candidate, Department of Mechanical Engineering, \href{mailto:marcon@mit.edu}{marcon@mit.edu}}, 
    Enrico M. Zucchelli\footnote{Postdoctoral Associate,  Department of Aeronautics and Astronautics, \href{mailto:ezucch@mit.edu}{enricoz@mit.edu}}, 
    Amit Jain\footnote{Postdoctoral Associate,  Department of Aeronautics and Astronautics, \href{mailto:ajain@mit.edu}{amitjain@mit.edu}}
}
\affil{Massachusetts Institute of Technology, Cambridge, Massachusetts 02139, USA}

\author{Victor Rodriguez-Fernandez\footnote{Associate Professor, Department of Computer Systems Engineering, \href{mailto:victor.rfernandez@upm.es}{victor.rfernandez@upm.es}}}
\affil{Universidad Politécnica de Madrid, Madrid 28038, Spain}

\author{Richard Linares.\footnote{Associate Professor, Department of Aeronautics and Astronautics, \href{mailto:linaresr@mit.edu}{linaresr@mit.edu} Senior Member AIAA.}}
\affil{Massachusetts Institute of Technology, Cambridge, Massachusetts 02139, USA}

\begin{document}

\maketitle

\begin{abstract}
This paper explores the application of Vision-Language Models (VLMs) as operator agents in the space domain, focusing on both software and hardware operational paradigms. Building on advances in Large Language Models (LLMs) and their multimodal extensions, we investigate how VLMs can enhance autonomous control and decision-making in space missions. In the software context, we employ VLMs within the Kerbal Space Program Differential Games (KSPDG) simulation environment, enabling the agent to interpret visual screenshots of the graphical user interface to perform complex orbital maneuvers. In the hardware context, we integrate VLMs with robotic systems equipped with cameras to inspect and diagnose physical space objects, such as satellites. Our results demonstrate that VLMs can effectively process visual and textual data to generate contextually appropriate actions, competing with traditional methods and non-multimodal LLMs in simulation tasks, and showing promise in real-world applications. 
\end{abstract}

\section{Introduction}
% General introduction, LLMs, Multimodal
\lettrine{L}{a}rge Language Models (LLMs) possess extensive domain-specific knowledge derived from their extensive pre-training, establishing themselves as invaluable tools across various fields. Since the emergence of the LLM trend, initiated with the first release of ChatGPT \cite{chatgpt}, these systems have undergone continuous development and have evolved into multimodal architectures. Multimodal models, such as GPT-4o \cite{openai2024gpt4o}, LLaMA 3.2 \cite{meta2024llama3-2} and Claude with its latest 3.5 Sonnet model \cite{anthropic_claude3_2024}, integrate language understanding with non-language capabilities, including vision and audio processing. This progression unlocks new opportunities for developing intelligent agents capable of recognizing and interpreting patterns not only at a semantic level but also through components that can incorporate other types of unstructured data into prompts, significantly expanding their potential applications and impact.

%VLMs and SOAT robotics
Extending these capabilities, Vision-Language Models (VLMs) build on multimodal principles by integrating visual reasoning into the LLM framework. By introducing new tokens into the prompts to process image frames, VLMs enable simultaneous semantic and visual reasoning. This enhancement is particularly valuable in dynamic applications like robotics, where the integration of vision and language reasoning enables systems to generate environment-responsive actions. Such actions, often described as descriptive policies, translate reasoning into meaningful, executable commands. Language models able to generate such commands are usually referred to as ``agentic". In particular, models such as OpenVLA \cite{kim2024openvlaopensourcevisionlanguageactionmodel} further advance this paradigm by incorporating two specialized tokenizers: one for spatial reasoning and another for characteristic reasoning, solidifying VLMs as the cornerstone of interactive and adaptable systems \cite{li2024visionlanguagefoundationmodelseffective}.

% Agents, VLM agents
The transformation of LLMs into highly capable LLM-based agents (LLMAs) marks a pivotal step toward achieving more human-like behavior in autonomous systems \cite{xi2023risepotentiallargelanguage}. A key proving ground for these agents lies in video games, which simulate complex and dynamic open-world environments, offering an ideal testbed for interaction and evaluation \cite{wang2023jarvis1openworldmultitaskagents}, \cite{tan2024cradleempoweringfoundationagents}, \cite{al2024projectsidmanyagentsimulations}. Agentic VLMs have emerged recently, and the field still holds significant potential for growth and development. Balrog \cite{paglieri2024balrogbenchmarkingagenticllm} is a benchmark designed to evaluate the reasoning capabilities of agentic LLMs and VLMs in gaming environments. While agentic VLMs have made a notable impact, they continue to underperform compared to LLM-based agents, highlighting the need for further refinement in visual reasoning.

% Our research
This research builds on previous advances in LLM-driven space control \cite{rodriguezfernandez2024languagemodelsspacecraftoperators}, \cite{2024sais.conf..247C}, \cite{zucchelienrico2024finetunedlanguagemodelsasspacesystemscontrollers} and explores the potential of vision-language models for space tasks \cite{foutter2024adaptingfoundationmodelspacebased}. We propose employing these models in the space domain, emphasizing an end-to-end space control framework that addresses two distinct operational paradigms: (1)~a software-based spacecraft control operator utilizing screenshots of the graphical user interface of a simulation environment, namely Kerbal Space Program, and (2)~a hardware-oriented robotic control system for inspecting physical space objects like satellites. Both paradigms align with common mission objectives, such as in-orbit servicing and satellite maintenance. 

This paper is organized as follows: Sec.~\ref{sec:backgrounds} provides a basic background on the applications in the space domain explored in our experiments. Sections~\ref{sec:VLMsoftware} and \ref{sec:VLMhardware} present two approaches: the first shows LLM capabilities in a software environment, while the second applies them to a real-world tasks. Given the preliminary nature of this paper, Secs.~\ref{sec:perspectives} and~\ref{sec:conclusions} provide an overview of our experimental setup, insights, and future steps.

\section{Background}
\label{sec:backgrounds}

This section provides the necessary background on the foundational technologies and tools relevant to this work, including large language models (LLMs), vision-language models (VLMs), and the Kerbal Space Program Differential Games (KSPDG).

\subsection{Large Language Models (LLMs)}
\label{subsec:llms}

Large Language Models (LLMs) are advanced neural networks trained on extensive text corpora. During pre-training, LLMs are taught to complete sentences. After that, with limited fine-tuning, LLMs can be aligned to perform a wide range of natural language processing (NLP) tasks. These models, typically based on transformer architectures, have demonstrated remarkable capabilities in tasks such as reasoning, summarization, translation, and autonomous decision-making. The scalability of LLMs, coupled with their ability to generalize knowledge across diverse domains, has rendered them indispensable for applications in autonomous systems. In these circumstances, LLMs can efficiently process complex inputs and produce coherent and interpretable output.

The versatility of LLMs is often unlocked through prompt engineering, a technique that structures model inputs to elicit desired behaviors. By carefully designing prompts, users can guide LLMs to perform complex reasoning tasks or exhibit high levels of contextual understanding without requiring explicit additional training.

\subsubsection{Prompt Engineering}
\label{subsubsec:prompt_engineering}

Prompt engineering is a key methodology for leveraging the capabilities of LLMs. It involves crafting input prompts to guide the model's behavior effectively. This approach is particularly powerful in scenarios where models are deployed in dynamic or resource-constrained environments, allowing users to achieve task-specific outcomes with minimal or no additional training. Two prominent paradigms in prompt engineering are \emph{zero-shot} and \emph{few-shot} prompting.

\paragraph{Zero-shot Prompting}
Zero-shot prompting involves presenting an LLM with a query or task description without providing any examples. The model relies on its extensive training to infer the appropriate response. For example, a zero-shot prompt for translation might look like:

\begin{verbatim}
    Translate the following sentence into French: "The weather is lovely today".
\end{verbatim}

This approach is highly efficient for general-purpose tasks, as it does not require additional context or fine-tuning.

\paragraph{Few-shot Prompting}
Few-shot prompting provides the model with a small number of examples within the prompt itself to establish context. This method enhances task-specific performance by grounding the model's output in the examples provided. For instance, a few-shot prompt for arithmetic reasoning could be structured as:

\begin{verbatim}
        Q: What is 3 + 5?
        A: 8
        Q: What is 7 + 9?
        A: 16
        Q: What is 12 + 15?
\end{verbatim}

Here, the examples serve as a guide for the model's response patterns, significantly improving accuracy in specialized domains.

\paragraph{Advanced Prompting Paradigms}
To further enhance reasoning capabilities, advanced prompting paradigms such as \emph{Chain of Thought} (CoT) \cite{wei2023chainofthoughtpromptingelicitsreasoning} and \emph{ReAct} (Reasoning and Acting) \cite{yao2023reactsynergizingreasoningacting} have been developed:

\begin{itemize}
    \item \textbf{Chain of Thought (CoT):} This paradigm encourages the model to produce intermediate reasoning steps before arriving at the final output. By breaking down complex problems into smaller, interpretable steps, CoT improves the model's accuracy and transparency in reasoning. For example, the following prompt can be included before the user input to activate CoT in the language model:

    \begin{verbatim}
    Q: If a train travels 50 miles per hour for 3 hours, how far does it travel? 
    Let's think step by step. 
    A: First, calculate the distance per hour: 50 miles/hour. 
    Next, multiply by the time traveled: 50 * 3 = 150 miles. 
    Therefore, the train travels 150 miles.
    \end{verbatim}

    \item \textbf{ReAct (Reasoning and Acting):} ReAct combines reasoning with decision-making to enable LLMs to handle interactive or agent-based tasks. The ReAct paradigm integrates logical deductions with contextual actions, making it suitable for autonomous systems and complex problem-solving. For example, the following prompt can be included before the user input to activate ReAct in the language model:

    \begin{verbatim}
    Q: The room is dark, and I need to find a flashlight. What should I do? 
    A: Reasoning: A flashlight is often stored in a drawer or cabinet.
       Action: Search the drawer for a flashlight.
    \end{verbatim}
\end{itemize}

These paradigms enhance the utility of LLMs in scenarios requiring high-skilled reasoning or agent-driven applications, making them integral to the advancement of AI-powered autonomous systems.

\paragraph{Structured Outputs}
Modern LLMs support structured outputs, including JSON and other machine-readable formats, making them well-suited for programmatic applications. These capabilities allow users to guide the model in generating outputs that integrate seamlessly with APIs or software systems.

For example:

\begin{verbatim}
Task: Provide a structured JSON response for the following user information:
Name: Alice, Age: 30, Location: New York.

    Output:
    {
        "name": "Alice",
        "age": 30,
        "location": "New York"
    }
\end{verbatim}

By clearly specifying the desired structure, users can ensure that the model's outputs are directly usable for downstream processes.

\paragraph{Function Calling}
Function calling builds on the structured JSON paradigm by enabling language models to interact with external codebases in an API-like manner. By specifying the desired tool or function in a structured JSON format, users can instruct the model to execute specific tasks or access external functionalities. For instance, a function call to retrieve weather information might look like this:

\begin{verbatim}
Task: Retrieve weather data for London using the function call.

    Output:
    {
        "function": "getWeatherData",
        "parameters": {
            "location": "London"
        }
    }
\end{verbatim}

\subsection{Vision-Language Models (VLMs)}
\label{subsec:vlms}

Vision-Language Models (VLMs) extend the capabilities ofLLMs by integrating visual information. By aligning textual and visual representations in a shared embedding space, VLMs can perform tasks that require understanding of both modalities. This multimodal capability makes them particularly suitable for simulation environments, where textual observations can be augmented with visual snapshots to improve situational awareness and decision making.

Recent advances in VLMs have connected the gap between computer vision and NLP. These models typically combine pre-trained vision encoders with language models to create powerful multimodal architectures. VLMs gained significant attention in 2021 with the introduction of CLIP~\cite{radford2021learningtransferablevisualmodels}, which pioneered contrastive learning for aligning vision and language modalities, enabling zero-shot transfer to a range of visual tasks. As visual-language models matured, frameworks like Flamingo~\cite{alayrac2022flamingovisuallanguagemodel} and BLIP~\cite{li2022blipbootstrappinglanguageimagepretraining} refined multimodal interactions, further improving the merge between vision and language models. More recently, LLaVA~\cite{liu2023visualinstructiontuning} showcased how instruction-following language models, integrated with CLIP-based visual encoders, could effectively handle complex multimodal tasks. Today, multimodal LLMs, such as GPT-4o \cite{openai2024gpt4o} and Claude 3.5 Sonnet \cite{anthropic_claude3_2024}, demonstrate unparalleled versatility, seamlessly integrating vision and language capabilities to serve as VLMs while excelling across diverse applications.

Once VLMs were established, their adaptation to robotics opened new frontiers. RoboFlamingo, for example, leverages pre-trained VLMs like OpenFlamingo, introducing fine-tuning strategies for effective language-conditioned robot control~\cite{li2024visionlanguagefoundationmodelseffective}. Similarly, OpenVLA~\cite{kim2024openvlaopensourcevisionlanguageactionmodel} combines advanced visual encoders and LLMs to achieve high-performance robot manipulation, highlighting the potential of multimodal systems in robotics. These advancements underscore how vision-language models are reshaping the landscape of autonomous systems and robotics.

VLMs have shown remarkable capabilities in various tasks, including:

\begin{itemize}[left=4em]
    \item Zero-shot image classification
    \item Image-text retrieval
    \item Visual question answering
    \item Object localization
    \item Multimodal reasoning
\end{itemize}

The architecture of modern VLMs often involves:

\begin{enumerate}[left=4em]
    \item A pretrained vision encoder (\textit{e.g.}, CLIP, DINOv2, SigLIP)
    \item A pretrained language model (\textit{e.g.}, Llama 2, Vicuna)
    \item A mechanism to align visual and textual representations (\textit{e.g.}, linear projections, attention mechanisms)
\end{enumerate}

By leveraging large-scale pretraining on diverse datasets, these models can generalize to a wide range of vision-language tasks, making them valuable tools for multimodal artificial intelligence~(AI) applications, including simulations and robotics \cite{wang2024solving}.

It is crucial to ensure that these tasks cannot be performed effectively by an LLM. As demonstrated in \cite{paglieri2024balrogbenchmarkingagenticllm}, LLMs continue to outperform VLMs when they can comprehend their reward structure or scoring criteria.

\subsection{Kerbal Space Program Differential Games (KSPDG)}
\label{subsec:kspdg}

Kerbal Space Program Differential Games(KSPDG)~\cite{allen2023spacegym} is a simulation environment based on the Kerbal Space Program~(KSP), developed by MIT Lincoln Laboratories as part of the space-gym library. Designed for reinforcement learning and differential game challenges, it offers a dynamic platform to navigate complex scenarios.

In KSPDG, agents operate in discrete steps, interpreting observations, such as relative positions, velocities, and fuel levels, and executing 4D actions: thrusts along the X-, Y-, and Z-axes, plus thrust impulse duration. Performance is evaluated on metrics such as approach time and distances to targets and obstacles. This process promotes adaptive decision making, enabling agents to optimize maneuvers in real time.

KSPDG 2025 features two main scenario categories: Pursuer-Evader and Target Guarding (Lady-Bandit-Guard), detailed below.

\subsubsection{Pursuer-Evader Scenarios}
\label{subsubsec:pursuer_evader}

In the Pursuer-Evader scenarios, participants design autonomous agents to control a pursuer spacecraft to minimize the distance to an evading spacecraft. The scenarios differ in the evader's strategies:

\begin{itemize}
    \item \textbf{E1:} No evasive maneuvers by the evader.
    \item \textbf{E2:} Random short-duration evasive maneuvers when the pursuer is within range.
    \item \textbf{E3:} Structured full-thrust maneuvers to escape within a distance threshold.
    \item \textbf{E4:} Constant prograde thrust aligned with the orbital velocity vector.
\end{itemize}

\subsubsection{Lady-Bandit-Guard Scenarios}
\label{subsubsec:lbg}

The Lady-Bandit-Guard (LBG) scenarios are 1-v-2 target guarding problems. Participants control the Bandit spacecraft, which aims to minimize the distance to the Lady spacecraft while maximizing the distance to the Guard spacecraft.

Scenarios vary in Lady and Guard policies as well as initial orbital configurations:

\begin{itemize}
    \item \textbf{Policy Environment Identifiers~(\textbf{lg}):}
    \begin{itemize}
        \item \textbf{lg0}: Lady and Guard are passive.
        \item \textbf{lg1}: Passive Lady, Guard pursues Bandit using a heuristic target-zero\_vel-target maneuver.
        \item \textbf{lg2}: Lady evades Bandit with out-of-plane burns, Guard pursues using heuristic target-zero\_vel-target maneuvers.
        \item \textbf{lg3}: Code-obfuscated environment, passive Lady, advanced Guard algorithms.
    \end{itemize}
    \item \textbf{Initial Orbit Environment Identifiers~(\textbf{i}):}
    \begin{itemize}
        \item \textbf{i1}: Lady and Guard in a circular orbit, Guard ~600m prograde of Lady, Bandit in an elliptical orbit with an upcoming tangential conjunction at apoapsis.
        \item \textbf{i2}: Lady, Bandit, and Guard in the same circular orbit, Guard ~600m retrograde of Lady, Bandit ~2000m retrograde of Guard.
    \end{itemize}
\end{itemize}

These scenarios are assessed using a fine-grained scoring equation defined by:

\begin{equation}
    \text{score} = \text{dm\_lb}^2 + \frac{a}{\text{dm\_bg} + b}
\end{equation}

where:
\begin{itemize}
    \item $\text{dm\_lb}$: Closest approach distance between the agent and the Lady (meters).
    \item $\text{dm\_bg}$: Closest approach distance between the agent and the Guard (meters).
    \item $a = 10^6$: A scaling factor ensuring that approximately 100 meters of $\text{dm\_lb}$ is as beneficial as 100 meters of $\text{dm\_bg}$ is harmful.
    \item $b = 0.1$: A positive offset to prevent division by zero and to balance the penalties and rewards appropriately.
\end{itemize}

\section{VLMs as Software operators - use case in Kerbal Space Program}
\label{sec:VLMsoftware}

To explore a distant spacecraft, satellite, or orbital debris, an agent must match its orbit to that of its target by executing a complex set of orbit correction maneuvers to reduce their relative position and velocity to zero. This process of approaching a target is commonly referred to as rendezvous. Once the agent achieves proximity within a sufficiently close range, it can transition to preparation for docking with the target to carry out the intended task. These operations demand precise trajectory planning and real-time adjustments to account for orbital mechanics, ensuring a safe and efficient approach.

\begin{figure}[ht]
\centering
\includegraphics[width=\textwidth]{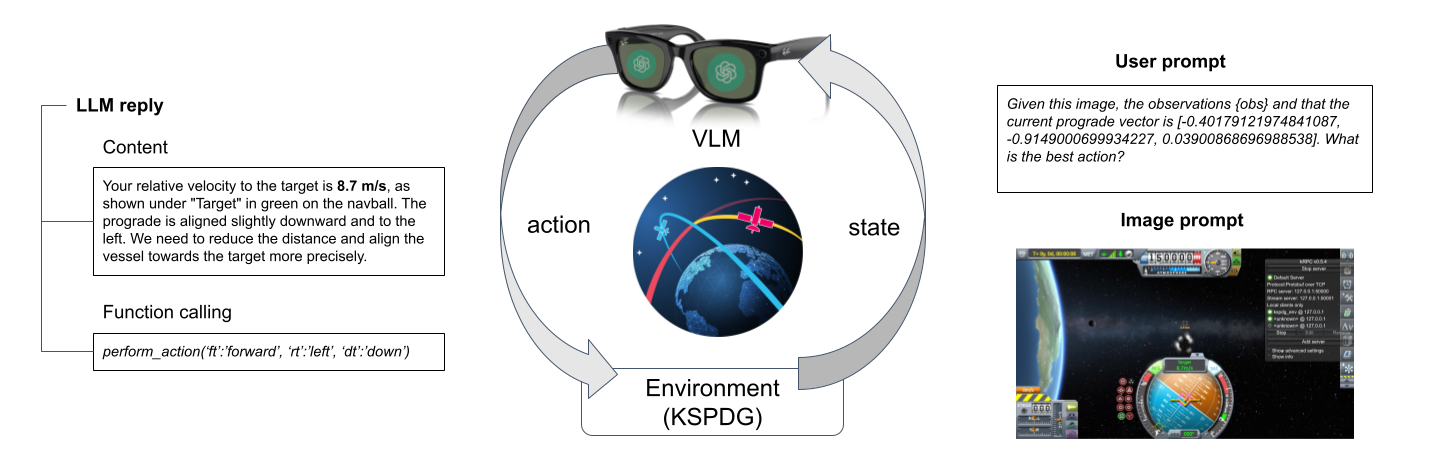}
\caption{Overview of the proposed approach to use an LLM (e.g., ChatGPT) as an autonomous spacecraft operator, with image prompts derived from in-game screenshots.}
\label{fig:ksp_llm_agent_vision}
\end{figure}

Figure~\ref{fig:ksp_llm_agent_vision} depicts an overview diagram of our new VLM agent interaction. In the previous challenge edition, an LLM was designed to only utilize text user prompts in real time leveraged with a fine-grained system prompt. The resulting content was aligned with few-shot examples, generating a concise response and providing function calls in the proper format. For this year's challenge, we extended the workflow by assigning specific observations to either the vision or language components of the VLM, allowing visual data to be processed through the vision module and textual data through the language module, leveraging each component's strengths.

This section focuses on our agentic approach within the simulation environment provided by the Kerbal Space Program Differential Games (KSPDG) gym environment, developed by MIT Lincoln Lab\cite{allen2023spacegym}. In our previous research centered on LLMs \cite{rodriguezfernandez2024languagemodelsspacecraftoperators} \cite{2024sais.conf..247C}, we observed significant difficulties for LLMs in interpreting the vessel's reference frame and effectively utilizing raw observational data presented as floating-point numbers. These challenges limited their ability to perform complex orbital maneuvers despite achieving surprising results mostly due to strong prompt engineering or model fine-tuning. To address these limitations, we are now exploring VLMs, which leverage the information presented in the in-game dashboard and provide all relevant data intuitively.

\subsection{Prior Prompt Engineering}
Our new agent retains similarities to its predecessor, which relied solely on language for operation \cite{rodriguezfernandez2024languagemodelsspacecraftoperators}. The original KSPDG action space uses continuous numbers, which many models find challenging to interpret effectively \cite{akhtar2023exploringnumericalreasoningcapabilities}, and these values were discretized in the same manner as in our approach. Similarly, the observation space, composed of distances and velocities, also relies on floating point numerical data, posing additional challenges for LLMs. To address these limitations, we augmented the observations with the prograde vector, as will be explained at the end of this subsection.
Additionally, models specialized in mathematical reasoning, such as OpenAI's o1\cite{openai2024o1}, exhibit latency levels that are impractical for the real-time missions addressed in this research. We discretize the action space into 9 distinct actions, encompassing various combinations of throttle levels and directional adjustments. Specifically, we include two throttle settings — full throttle and full reverse throttle - for each axis of motion (\textit{X}, \textit{Y}, and \textit{Z}), as well as a no-throttle option.

The following actions represent movement along the three axes of the spacecraft's reference frame:
\begin{itemize}
    \item \textbf{X-axis:} Left and right adjustments.
    \item \textbf{Y-axis:} Forward and backward adjustments.
    \item \textbf{Z-axis:} Up and down adjustments.
\end{itemize}

Each axis allows for three possible actions: positive thrust, negative thrust, or no thrust. This configuration results in a total of $3 \times 3 \times 3 = 27$ possible permutations of combined actions, enabling the agent to freely move in the 3D space.

The previous prompting strategy emphasized providing the agent with explicit telemetry details and a contextual mission goal from the observation space. It included structured information, detailed below:
\begin{itemize}
    \item Relative positions and velocities between the spacecraft, the target, and optionally, the guard to evade.
    \item Spacecraft status such as fuel usage.
    \item Augmented observations (not included by default in the ones given by the KSPDG challenge organizers) including the prograde vector.
\end{itemize}

While effective in many scenarios, relying uniquely on textual prompts constrained the previous agent’s capacity to comprehend the spatial environment, especially in dynamically evolving situations. To address this, we incorporate visual inputs and redesign the prompt engineering process, introducing examples that explicitly describe the game’s dashboard graphic user interface~(GUI) and most important mission metrics. Our previous research included several new metrics derived from data augmentation. Due to the vision capabilities, this study only augments observations by providing the prograde, an important component for mission control.

The prograde $\mathbf{p}$ can be calculated by:

\begin{equation}
\mathbf{p} = \mathbf{R}^{-1} \frac{\mathbf{v}_p - \mathbf{v}_e}{\|\mathbf{v}_p - \mathbf{v}_e\|}
\end{equation}

where:

\begin{itemize}
\item $\mathbf{R}$: Rotation matrix that transforms coordinates from vessel to celestial body reference frame.
\item $\mathbf{v}_p, \mathbf{v}_e$: pursuer's and evader's velocity in celestial body reference frame.
\end{itemize}

\subsection{Vision Few-Shot Prompting}
Both system and user prompts are crucial for the model's performance, guiding the LLM to interpret the interface much the same way a human relies on a manual for reference. By incorporating visual data through in-game screenshots, the vision agent can dynamically observe key elements that are essential to completing the task. These visual clues enhance the textual inputs, allowing the agent to form a deeper understanding of the mission state and respond more effectively to changing conditions.

\begin{figure}[ht]
\centering
\fbox{
\begin{minipage}{\textwidth}
    \begin{minipage}{0.4\textwidth}
        \centering
        \includegraphics[width=\textwidth,keepaspectratio]{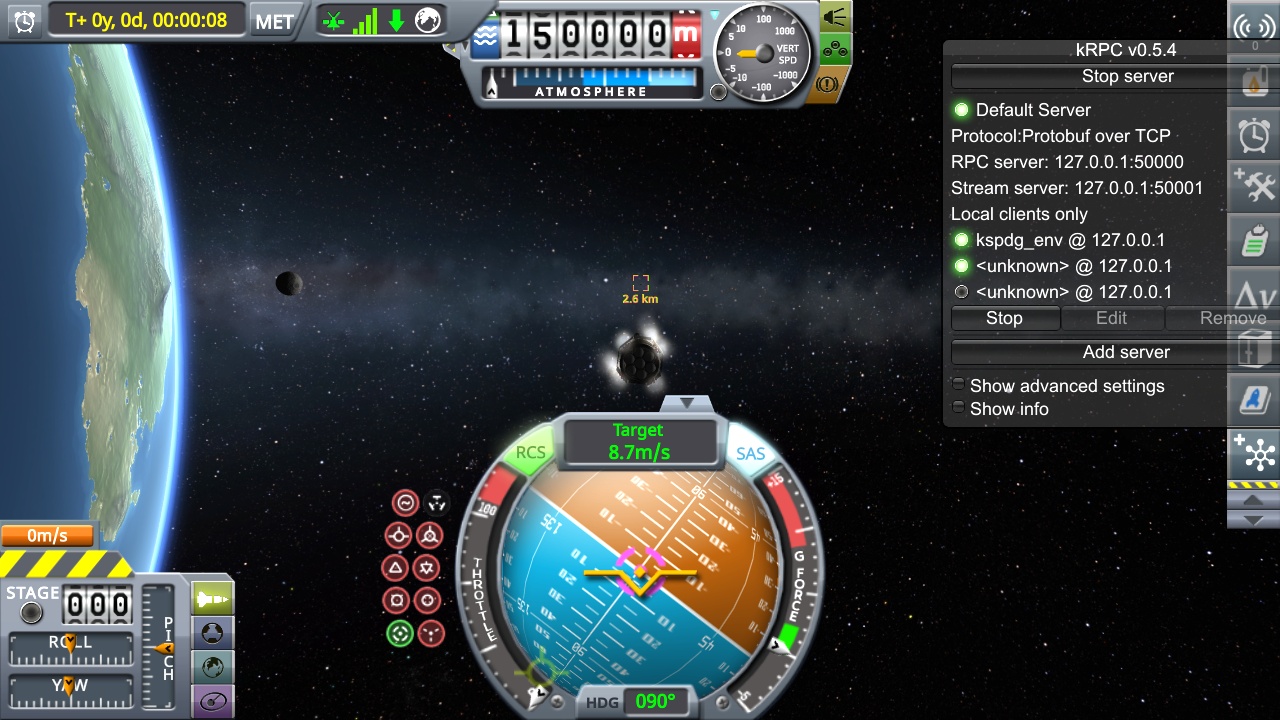}
        \caption*{Visual Input}
    \end{minipage}%
    \hfill
    \begin{minipage}{0.55\textwidth}
        \texttt{\textbf{Input:}\\
        Current distance to Lady: 2.6~kilometers; Prograde far in the bottom left side of the navball; Current speed: 8.7~m/s.\\
        \textbf{Reasoning:}\\
        Apply Forward throttle to start our approach, adjust prograde using Right and Up throttles.\\
        \textbf{Output:}\\
        perform\_action(Forward Throttle: Forward, Right Throttle: Right, Down Throttle: Up)
        }
    \end{minipage}
\end{minipage}
}
\caption{Few-shot example illustrating the input, reasoning, and output for controlling a spacecraft.}
\label{fig:few_shot_example}
\end{figure}

The few-shot examples in Fig.~\ref{fig:few_shot_example} demonstrate how to process and act upon this visual information in conjunction with telemetry. These example demonstrations highlight:
\begin{itemize}
    \item How to interpret the navball to align with the prograde or retrograde vectors.
    \item How to use the visual representation of the spacecraft's orientation to adjust maneuvers.
    \item How to combine visual screenshots with textual telemetry to generate accurate action commands.
\end{itemize}

The actual telemetry read by the model receives less observations than our previous agent from \cite{rodriguezfernandez2024languagemodelsspacecraftoperators}, reading only relative positions and prograde values collected through data augmentation.

Additionally, visual input processing involves extracting key features, such as visual mission markers displayed on the KSP navball. These features are ingested into the model's prompt alongside textual telemetry for multimodal reasoning. The said approach provides a reliable framework for guiding the model towards more generalizable reasoning, leveraging the dashboard's comprehensive data.

It should be noted that, while VLMs accurately recognize various visual cues, some metrics cannot be intuitively interpreted through visual means. For instance, it cannot visually interpret motion dynamics, such as inertia or proximity changes, under the current state of the research.

\subsection{Results}

As mentioned earlier, this paper is a preliminary development and evaluation of our proposed agents. Therefore, this section only includes a simple approach metric, the distance from the Lady vessel as well as the distance to the Guard spacecraft, which the Bandit, our spacecraft, must avoid in this mission. Finally, we present the scoring metric defined in Section~\ref{subsubsec:lbg} along with the latency metrics for each VLM model.

As shown in Table \ref{tab:KSP_Table_Results}, both the LLM and the VLM agents\footnote{The VLM agents used in this study include Claude 3.5 (Anthropic), GPT-4.0 (OpenAI), and LLaMA 3.2 (Meta).} outperform most traditional methods\footnote{Results extracted from the SpaceGym paper by Ross Allen \cite{allen2023spacegym}}. Furthermore, we observe that LLMs demonstrate strong reasoning capabilities, as reflected in their ability to balance the competing objectives of minimizing the distance to the Lady while maintaining a safe distance from the Guard. 

This scoring function rewards approaches that minimize $\text{dm\_lb}$ while penalizing those that reduce $\text{dm\_bg}$ excessively. As such, it reflects the dual objectives of the task: approaching the Lady while keeping a safe distance from the Guard.

\begin{table}[h]
\centering
\small % Reduce font size
\setlength{\tabcolsep}{19pt} % Adjust column spacing
\begin{tabular}{l|ccc|c}
\hline \hline
\textbf{Method} & \textbf{Best Dist. (m)} & \textbf{Avg. Dist. (m)} & \textbf{Avg. to Guard (m)} & \textbf{Avg. Score} \\ \hline \hline
Naive & 225.0 & 225.0 & - & - \\
PPO & >1643 & 2346 & - & - \\
iLQGames & >53.31 & 60.99 & - & - \\
Lambert-MPC & >18.24 & 47.94 & - & - \\
\hline
ChatGPT LLM CoT & 317.63 & 487.58 & \underline{360.74} & 260,901 \\
Claude LLM CoT & 48.22 & 79.18 & 89.59 & \dashuline{18,938} \\
LLaMA LLM CoT & 1708.78 & 1956.92 & \textbf{1392.82} & 3,901,235 \\
\hline
LLaMA LLM FT & \textbf{3.62} & \textbf{12.98} & 12.07 & 105,045 \\ \hline
ChatGPT VLM & 15.74 & \dashuline{30.42} & 234.30 & \underline{5,626} \\
Claude VLM & \dashuline{12.68} & 52.07 & 174.58 & 21,488 \\
LLaMA VLM & \underline{4.76} & \underline{18.36} & \dashuline{291.33} & \textbf{3,961} \\
\hline \hline
\end{tabular}
\caption{Performance metrics for various models and techniques. CoT refers to Chain of Thought reasoning (applied to the rendezvous problem), and FT refers to Fine-Tuning. \textbf{Bold} indicates the best performance, \underline{underline} the second best, and \dashuline{dashed underline} the third best..}
\label{tab:KSP_Table_Results}
\end{table}

% ADD LATENCIES
\begin{table}[h]
\centering
\small % Reduce font size
\setlength{\tabcolsep}{15pt} % Adjust column spacing
\begin{tabular}{l|ccc}
\hline \hline
\textbf{VLM Model} & \textbf{Best Latency (ms)} & \textbf{Worst Latency (ms)} & \textbf{Average Latency (ms)} \\ \hline \hline
ChatGPT VLM & \textbf{2982} & \textbf{7120} & \textbf{4756} \\
Claude VLM & 6129 & 7409 & 9373 \\
LLaMA VLM 3 & 7139 & 9664 & 11317 \\
\hline \hline
\end{tabular}
\caption{Latency comparison for Visual Language Models (VLMs).}
\label{tab:VLM_Latency}
\end{table}

Fine-tuned LLMs remain effective in approaching the Lady, showcasing strong performance in proximity metrics. However, VLMs achieve better overall scores, indicating their ability to generalize across objectives. In particular, VLMs deliver solid results in adapting to the guard evasion problem, leveraging spatial and contextual cues from visual input. In contrast, the LLMs tested in this study were guided by more specialized prompts for the rendezvous problem, which likely influenced their ability to balance proximity and guard avoidance. However, the Balrog benchmark \cite{paglieri2024balrogbenchmarkingagenticllm} suggests that with refined instructions or task-specific adjustments, LLMs achieve performance superior to VLMs in the current state of these models.

We also observe in Table~\ref{tab:VLM_Latency} that the primary bottleneck in these models is their latency. Therefore, it is crucial to carefully consider the implementation of agents to account for these high latencies.

Although further refinement of novel VLM models is needed, as highlighted in the Balrog benchmark, it is intuitive to anticipate that VLMs will generalize more effectively in such tasks as they continue to mature, though this capability remains limited at present. We showcased that these models interpret spatial and contextual cues, which enables them to balance proximity to the Lady and avoidance of the Guard, aligning well with the scoring function.

This area remains ripe for exploration, with future advancements in fine-tuning or hybrid approaches offering the potential to enhance the performance of both LLMs and VLMs while preserving their generalization capabilities across diverse tasks and environments.

\section{VLMs as Hardware operators - Robotics use case}
\label{sec:VLMhardware}
This section explores the integration of Vision-Language Models (VLMs) with robotic systems, focusing on their application in space hardware inspection. By leveraging real-time image processing and model-driven decision-making, the system aims to enhance autonomous robotics for complex tasks in dynamic environments. The following details the interaction between the robot, its camera, and the VLM in diagnosing potential issues within space hardware.

\subsection{VLM - Robot Interaction}
Historically, robots have been highly specialized, excelling at narrowly defined tasks such as assembly line work or material handling. While those robots are effective within their specific domains, they lack the versatility to handle a wide range of activities. Typically, each new task would require the development of a separate robot, creating inefficiencies and increasing the need for specialized infrastructure. However, a growing trend in the robotics industry is the shift toward more generalized robots—like humanoid forms—that can adapt to a broad spectrum of tasks. This trend has been particularly evident in industries such as automotive manufacturing, where humanoid robots are now used not only for repetitive tasks like welding but also for more complex operations such as quality inspections and maintenance. The ability to use the same tools as human workers and switch between diverse tasks reduces the need for multiple specialized machines, making operations more efficient and cost-effective.

The proposed system employs an xArm 7 robotic arm~\cite{xArm7inproceedings} equipped with an Intel RealSense camera~\cite{keselman2017intelrealsensestereoscopicdepth} to inspect and diagnose potential issues with space hardware. The hardware setup allows the robot to capture red, green, blue, depth~(RGBD) images and stream them as frames to the VLM, enabling real-time interaction between the robot and the environment. In the system that would ultimately be deployed, a LIDAR sensor would be used instead of this camera to achieve higher precision.

\begin{figure}[ht]
\centering
\includegraphics[width=\textwidth]{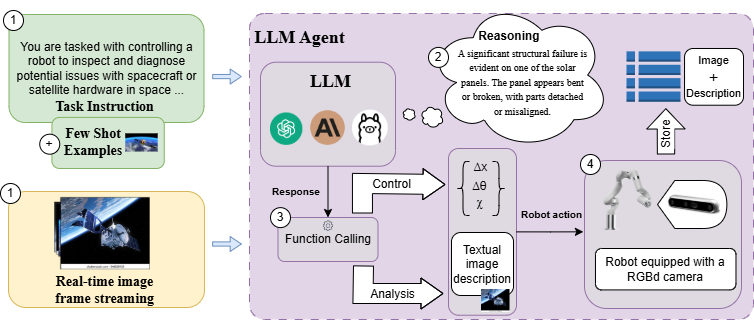}
\caption{Overview of the LLM-based robotic control system for space hardware inspection and diagnosis.}
\label{fig:robot_llm_agent_vision}
\end{figure}

The overall pipeline, illustrated in Fig.~\ref{fig:robot_llm_agent_vision}, operates by dividing the video stream from the RealSense camera into individual image frames, which are ingested by the VLM one at a time. Each frame is processed sequentially, and the model reasons about its content to determine the next control action. The generated action includes positional changes ($\Delta x$), rotational adjustments ($\Delta \theta$), and a boolean flag ($\chi$) to decide whether to capture an annotated image. This can be expressed mathematically as:
\begin{equation}
a_t = \mathcal{F}(I_t, S_t, P)
\end{equation}
where
\begin{equation}
a_t = (\Delta x_t, \Delta \theta_t, \chi_t)
\end{equation}
represents the action generated at timestep $t$:
\begin{itemize}
    \item $\Delta x_t$: Positional change,
    \item $\Delta \theta_t$: Rotational adjustment,
    \item $\chi_t$: Boolean flag indicating whether to capture an annotated image.
\end{itemize}
Here, $I_t$ denotes the input image frame at time~$t$, $S_t$ is the robot's corresponding state (e.g., Cartesian position and joint angles), and $P$ is the system prompt containing task-specific instructions. $\mathcal{F}$ represents the Vision-Language Model (VLM) that processes these inputs to generate the action. Importantly, the system ensures that no subsequent frames are processed until the current action is executed, maintaining a synchronized pipeline between perception and control.

Inference is the most time-consuming component of LLM processing. To address this limitation, we implement advanced optimization strategies, including refined prompting techniques and LLM-specific mechanisms such as Flash Attention \cite{dao2022flashattentionfastmemoryefficientexact} or open-source models; for proprietary models, we consider strategies to reduce the output prompt size via few-shot prompting examples or user-side prompt caching. These measures aim to significantly minimize response times, ensuring efficient and responsive performance for mission-critical real-time operations.

Once an action is completed, the robot updates its position and orientation, and the pipeline resumes processing the next frame. This iterative approach ensures structured decision making and efficient data collection, with images and associated descriptions stored for post-mission analysis.

\subsection{Methodology and Preliminary Results}

The latest robotic trends have led us to explore \textit{OpenVLA}~\cite{kim2024openvlaopensourcevisionlanguageactionmodel}, a VLA model that unifies ``controller'' and image capture decisions through a 7-dimensional action space (a 6D Cartesian vector plus the boolean snapshot decision, $\chi$). This model demonstrates great results compared to other VLMs and state-of-the-art diffusion models, and also offers a generalizable framework for several robotics tasks.

A model used for a complex scenario such as the one visualized in Table~\ref{fig:robot_llm_agent_vision} must be either trained or fine-tuned to ensure effectiveness, requiring domain-specific datasets with a variety of desired tasks. In this work, we fine-tune~\textit{OpenVLA} on a curated set of training samples, tracking standard performance metrics (\textit{e.g.}, accuracy and 
loss curves) to evaluate convergence. Furthermore, we develop a real-time 
teleoperation interface that records the robot's states - including Cartesian 
positions, joint angles, and efforts - via keystrokes. This setup serves both as a mechanism for data collection and a means for interactive validation of the 
fine-tuned model.

Our fine-tuning process uses an HDF5 dataset, a hierarchical format that organizes data into groups and datasets. Each session in the dataset represents a training episode with multiple frame groups. These frame groups include image data, robot states, and a single instruction for the entire episode. By shuffling the frames within each session, the model learns to predict actions for each timestep independently, avoiding reliance on the sequence of frames and improving its generalization.

Table~\ref{tab:loss_metrics} summarizes the performance metrics from a preliminary fine-tuning on a 90-10 split dataset with only 10 episodes, significantly fewer than the recommended 50 for effective training.

\begin{table}[h!]
\centering
\begin{tabular}{|c|c|c|c|c|c|}
\hline
\textbf{Metric} & \textbf{Training Loss} & \textbf{Training L1 Loss} & \textbf{Validation Loss} & \textbf{Validation L1 Loss} & \textbf{Batch Size} \\ \hline
\textbf{Experiment 1}  & 0.748                  & 0.054                     & 1.424                   & 0.067                      & 8                  \\ \hline
\textbf{Experiment 2}  & 0.461                  & 0.021                     & 1.237                   & 0.049                      & 16                  \\ \hline
\end{tabular}
\caption{Loss metrics for the \textit{OpenVLA} model fine-tuning experiments. Each row represents a different configuration.}
\label{tab:loss_metrics}
\end{table}

\begin{table}[h!]
\centering
\begin{tabular}{|c|c|c|}
\hline
\textbf{Metric} & \textbf{Training Accuracy (\%)} & \textbf{Validation Accuracy (\%)} \\ \hline
\textbf{Experiment 1} & 70                        & 58.9                             \\ \hline
\textbf{Experiment 2} & 83.4                        & 69.5                             \\ \hline
\end{tabular}
\caption{Accuracy metrics for the \textit{OpenVLA} model fine-tuning experiments.}
\label{tab:accuracy_metrics}
\end{table}

%% Need to talk about results

Although the training loss does not provide any apparent insight, its convergence and improvement with larger batch sizes suggest that the model benefits from the increased capacity, even within the constraints of the RTX 4090 hardware. On the other hand, the L1 loss, which calculates the deviation between the ground truth action and the predicted one, offers a clearer view of the performance of the model. With an average deviation of 50 millimeters in the final training steps, the results indicate reasonable accuracy at this stage. The model's performance, achieved with just one-fifth of the recommended dataset size, underscores its promising capabilities despite the evident need for further refinement. The improvement in accuracy to 83.4\% in Experiment 2 demonstrates measurable progress, while the relatively close validation accuracy ensures the model is not overfitting. These preliminary results suggest the model’s capacity to handle the new inspection task.

%% This concludes the section
The transition to more adaptable, multi-purpose robots reflects the increasing demand for systems that can evolve alongside changing environments and tasks. Humanoid robots, due to their ability to mimic human actions and interact with tools in a human-like manner, represent the next frontier in this evolution. These robots not only improve efficiency but also open up new use cases that would have previously required entirely different robotic systems. For instance, humanoid robots can assist in high-stakes environments like space exploration, where tasks often require both cognitive reasoning and physical dexterity. By exploring the capabilities of systems like the xArm 7, we move closer to a future where robots are not limited to rigid, predefined actions but can perform dynamic and complex tasks with the precision and flexibility of human workers. As robotic systems continue to evolve, particularly in the context of space exploration, the integration of VLMs will be key to unlocking the full potential of autonomous operations, allowing robots to adapt in real-time and collaborate seamlessly with human operators in environments that are too harsh, distant, or hazardous for human astronauts.
%This hardware-software pipeline showcases the potential of VLMs for autonomous robotic operations in space, particularly for tasks requiring visual reasoning and decision-making. However, this is a preliminary implementation, and challenges such as minimizing latency in frame processing, improving model accuracy for novel scenarios, and ensuring operational safety remain critical areas for future exploration. Further fine-tuning of the VLM and optimization of the hardware pipeline are expected to enhance performance in subsequent iterations.
% future plan
% 
%
%The xArm 7 serves as a starting point
%Use it to demonstrate the software's capability to identify, pick, and place objects
%Using a 7 degree of freedom robot for increased agility in performing tasks
%inspired by the capabilities of the human arm

%specialized robots excel at a specific, narrowly defined task like pick and place, material handling
%However, they are not generalizable - they perform poorly at other tasks
%each task would require an entirely separate robot.
%There has been a trend to more generalized robots. For example, several car manufacturer have moved to using humanoids in their assembly line to improve the efficiency of certain tasks, for instance to weld, perform inspections

%as they are able to use tools that humans can use
%they can perform more complex tasks, different use cases for the same robot

%exploring the capabilities of using humanoids

\section{Perspectives}
\label{sec:perspectives}

%
%this technology is very promising
%these recent advancements offer great 
%discussed thus far offer great potential for a variety of applications ranging from satellite %inspection and servicing to ===

%traditional approaches - deterministic 
%well-defined dynamics and kinematics
%well defined scenarios and mission objectives
%have been used with some success in limited scope scenarios (cooperative rendezvous)

%
%not suitable for more complex applications  - applications that are increasingly in demand
%cislunar
%looking to tackle increasingly difficult challenges like satellite inspection and servicing
%dynamic illumination (especially in low earth orbit)
%uncertain or unknown parameters (mass of the target spacecraft, moment of inertia)
%rapidly changing mission objectives - determining the best course of action online

%this versatile framework allows effective solutions to be deployed

%collaborative solutions - in-orbit assembly
%assisting a human operator

%initial tests performed on the UFactory xArm 7
%have considered a humanoid form factor - chosen because of versatility

% Si quieres separar ideas usa comments - ok ok

% draft
Recent advancements in Visual Language Models (VLMs) show great promise for transforming the way we operate spacecraft and robots in space. This emerging technology offers significant potential for a wide range of applications, from satellite inspection and servicing to in-orbit assembly and collaborative missions. Unlike traditional, deterministic approaches that rely on well-defined dynamics, kinematics, and mission objectives, VLMs provide a versatile framework capable of adapting to the complexities of real-world space environments. These environments are often characterized by uncertain parameters such as the mass or moment of inertia of target spacecraft, dynamic lighting conditions (especially in low Earth orbit), and rapidly changing mission objectives. VLMs are well-suited for tackling these challenges, as they allow for online decision-making and flexible responses to unanticipated scenarios.

Traditional approaches have had success in more limited, well-defined scenarios—such as cooperative rendezvous missions—where the spacecraft's actions and objectives are relatively well-defined. However, these methods are increasingly inadequate for more complex and dynamic tasks, such as those required for satellite inspection, servicing, and cislunar missions. In such cases, VLMs enable spacecraft and robotic systems to operate autonomously or with human assistance, dynamically adjusting to changing conditions and evolving mission goals. Initial tests, such as those performed on the UFactory xArm 7 and humanoid form factors, have shown promise in demonstrating the potential for this technology. The humanoid design is especially valuable for its versatility, providing a flexible platform for a variety of space-based tasks, from collaborative operations to assisting human operators in real-time. As this technology continues to develop, it is expected to play a crucial role in overcoming the increasingly difficult challenges of space exploration and operations.

\section{Conclusions}
\label{sec:conclusions}

% draft; I NEED MORE OF A PUNCHLINE. THIS RESULTS LET US TALK ABOUT THIS AND RECENT ADVANCEMENTS SUPPORT A BETTER NARRATIVE.

In conclusion, Large Language Models (LLMs) have evolved significantly from their initial form, becoming increasingly capable and versatile tools in a wide range of domains. The advent of multimodal models, such as GPT-4o and LLaMA 3.2, has expanded the ability of LLMs to process and integrate diverse types of unstructured data, enhancing their utility in complex tasks that require both language and non-language reasoning. This transformation is particularly evident in Vision-Language Models (VLMs), which integrate visual and linguistic processing to enable more sophisticated reasoning, particularly in dynamic environments like robotics. As these models continue to advance, they pave the way for intelligent agents capable of executing actions that are both contextually aware and adaptable.

The transition from LLMs to LLM-based agents (LLMAs) marks a significant leap toward creating more autonomous, human-like systems. Video games, with their complex and interactive environments, have become an important proving ground for these agentic models. However, despite the considerable progress in VLMs, challenges remain, particularly in visual reasoning capabilities, which still require refinement to reach the level of performance seen in purely language-based agents.

Our research builds on these foundational developments by exploring the application of LLMs and VLMs in the space domain, with a focus on space control tasks. We propose an innovative end-to-end space control framework that spans both software-based spacecraft control and hardware-oriented robotic inspection of space objects. This dual approach offers valuable insights into the potential of LLMs for both simulation and real-world applications in space exploration and satellite maintenance. As we move forward, our work contributes to the ongoing evolution of LLMs and their integration into autonomous systems across diverse fields, with the space domain serving as an exciting and challenging frontier for future research.

% \section*{Appendix}
% An Appendix, if needed, should appear before the acknowledgments.

\section*{Acknowledgments}
% TODO: UPM fellowship
% TODO: AIA Spaceforce

Research was sponsored by the UPM Fellowship and by the Department of the Air Force Artificial Intelligence Accelerator and was accomplished under Cooperative Agreement Number FA8750-19-2-1000. The views and conclusions contained in this document are those of the authors and should not be interpreted as representing the official policies, either expressed or implied, of the Department of the Air Force or the U.S. Government.

\bibliography{references}

\begin{thebibliography}{28}
\newcommand{\enquote}[1]{``#1''}
\providecommand{\natexlab}[1]{#1}
\providecommand{\url}[1]{\texttt{#1}}
\providecommand{\urlprefix}{URL }
\expandafter\ifx\csname urlstyle\endcsname\relax
  \providecommand{\doi}[1]{\discretionary{}{}{}https://doi.org/#1}\else
  \providecommand{\doi}[1]{\discretionary{}{}{}\urlstyle{rm}\url{https://doi.org/#1}}\fi

\bibitem[{OpenAI(2022)}]{chatgpt}
OpenAI, \enquote{Introducing ChatGPT,} \url{https://openai.com/blog/chatgpt}, 2022.
\newblock (Accessed: 11-27-2024).

\bibitem[{OpenAI(2024{\natexlab{a}})}]{openai2024gpt4o}
OpenAI, \enquote{Hello, GPT-4o!} , May 2024{\natexlab{a}}.
\newblock \urlprefix\url{https://openai.com/index/hello-gpt-4o/}, accessed: 11-28-2024.

\bibitem[{AI(2024)}]{meta2024llama3-2}
AI, M., \enquote{Llama 3.2: Vision and Edge for Mobile Devices,} , September 2024.
\newblock \urlprefix\url{https://ai.meta.com/blog/llama-3-2-connect-2024-vision-edge-mobile-devices/}, accessed: 2024-11-28.

\bibitem[{{Anthropic}(2024)}]{anthropic_claude3_2024}
{Anthropic}, \enquote{Introducing Claude 3.5: Faster, safer, smarter, and with a sense of humor,} , June 2024.
\newblock \urlprefix\url{https://www.anthropic.com/news/claude-3-5-sonnet}, accessed: 2024-12-02.

\bibitem[{Kim et~al.(2024)Kim, Pertsch, Karamcheti, Xiao, Balakrishna, Nair, Rafailov, Foster, Lam, Sanketi, Vuong, Kollar, Burchfiel, Tedrake, Sadigh, Levine, Liang, and Finn}]{kim2024openvlaopensourcevisionlanguageactionmodel}
Kim, M.~J., Pertsch, K., Karamcheti, S., Xiao, T., Balakrishna, A., Nair, S., Rafailov, R., Foster, E., Lam, G., Sanketi, P., Vuong, Q., Kollar, T., Burchfiel, B., Tedrake, R., Sadigh, D., Levine, S., Liang, P., and Finn, C., \enquote{OpenVLA: An Open-Source Vision-Language-Action Model,} , 2024.
\newblock \urlprefix\url{https://arxiv.org/abs/2406.09246}.

\bibitem[{Li et~al.(2024)Li, Liu, Zhang, Yu, Xu, Wu, Cheang, Jing, Zhang, Liu, Li, and Kong}]{li2024visionlanguagefoundationmodelseffective}
Li, X., Liu, M., Zhang, H., Yu, C., Xu, J., Wu, H., Cheang, C., Jing, Y., Zhang, W., Liu, H., Li, H., and Kong, T., \enquote{Vision-Language Foundation Models as Effective Robot Imitators,} , 2024.
\newblock \urlprefix\url{https://arxiv.org/abs/2311.01378}.

\bibitem[{Xi et~al.(2023)Xi, Chen, Guo, He, Ding, Hong, Zhang, Wang, Jin, Zhou, Zheng, Fan, Wang, Xiong, Zhou, Wang, Jiang, Zou, Liu, Yin, Dou, Weng, Cheng, Zhang, Qin, Zheng, Qiu, Huang, and Gui}]{xi2023risepotentiallargelanguage}
Xi, Z., Chen, W., Guo, X., He, W., Ding, Y., Hong, B., Zhang, M., Wang, J., Jin, S., Zhou, E., Zheng, R., Fan, X., Wang, X., Xiong, L., Zhou, Y., Wang, W., Jiang, C., Zou, Y., Liu, X., Yin, Z., Dou, S., Weng, R., Cheng, W., Zhang, Q., Qin, W., Zheng, Y., Qiu, X., Huang, X., and Gui, T., \enquote{The Rise and Potential of Large Language Model Based Agents: A Survey,} , 2023.
\newblock \urlprefix\url{https://arxiv.org/abs/2309.07864}.

\bibitem[{Wang et~al.(2023)Wang, Cai, Liu, Jin, Hou, Zhang, Lin, He, Zheng, Yang, Ma, and Liang}]{wang2023jarvis1openworldmultitaskagents}
Wang, Z., Cai, S., Liu, A., Jin, Y., Hou, J., Zhang, B., Lin, H., He, Z., Zheng, Z., Yang, Y., Ma, X., and Liang, Y., \enquote{JARVIS-1: Open-World Multi-task Agents with Memory-Augmented Multimodal Language Models,} , 2023.
\newblock \urlprefix\url{https://arxiv.org/abs/2311.05997}.

\bibitem[{Tan et~al.(2024)Tan, Zhang, Xu, Xia, Ding, Li, Zhou, Yue, Jiang, Li, An, Qin, Zong, Zheng, Wu, Chai, Bi, Xie, Gu, Li, Zhang, Tian, Wang, Wang, Karlsson, An, Yan, and Lu}]{tan2024cradleempoweringfoundationagents}
Tan, W., Zhang, W., Xu, X., Xia, H., Ding, Z., Li, B., Zhou, B., Yue, J., Jiang, J., Li, Y., An, R., Qin, M., Zong, C., Zheng, L., Wu, Y., Chai, X., Bi, Y., Xie, T., Gu, P., Li, X., Zhang, C., Tian, L., Wang, C., Wang, X., Karlsson, B.~F., An, B., Yan, S., and Lu, Z., \enquote{Cradle: Empowering Foundation Agents Towards General Computer Control,} , 2024.
\newblock \urlprefix\url{https://arxiv.org/abs/2403.03186}.

\bibitem[{AL et~al.(2024)AL, Ahn, Becker, Carroll, Christie, Cortes, Demirci, Du, Li, Luo, Wang, Willows, Yang, and Yang}]{al2024projectsidmanyagentsimulations}
AL, A., Ahn, A., Becker, N., Carroll, S., Christie, N., Cortes, M., Demirci, A., Du, M., Li, F., Luo, S., Wang, P.~Y., Willows, M., Yang, F., and Yang, G.~R., \enquote{Project Sid: Many-agent simulations toward AI civilization,} , 2024.
\newblock \urlprefix\url{https://arxiv.org/abs/2411.00114}.

\bibitem[{Paglieri et~al.(2024)Paglieri, Cupiał, Coward, Piterbarg, Wolczyk, Khan, Pignatelli, Łukasz Kuciński, Pinto, Fergus, Foerster, Parker-Holder, and Rocktäschel}]{paglieri2024balrogbenchmarkingagenticllm}
Paglieri, D., Cupiał, B., Coward, S., Piterbarg, U., Wolczyk, M., Khan, A., Pignatelli, E., Łukasz Kuciński, Pinto, L., Fergus, R., Foerster, J.~N., Parker-Holder, J., and Rocktäschel, T., \enquote{BALROG: Benchmarking Agentic LLM and VLM Reasoning On Games,} , 2024.
\newblock \urlprefix\url{https://arxiv.org/abs/2411.13543}.

\bibitem[{Rodriguez-Fernandez et~al.(2024)Rodriguez-Fernandez, Carrasco, Cheng, Scharf, Siew, and Linares}]{rodriguezfernandez2024languagemodelsspacecraftoperators}
Rodriguez-Fernandez, V., Carrasco, A., Cheng, J., Scharf, E., Siew, P.~M., and Linares, R., \enquote{Language Models are Spacecraft Operators,} , 2024.
\newblock \urlprefix\url{https://arxiv.org/abs/2404.00413}.

\bibitem[{{Carrasco} et~al.(2024){Carrasco}, {Rodriguez-Fernandez}, and {Linares}}]{2024sais.conf..247C}
{Carrasco}, A., {Rodriguez-Fernandez}, V., and {Linares}, R., \enquote{{Fine-tuning LLMs for Autonomous Spacecraft Control: A Case Study Using Kerbal Space Program},} \emph{Proceedings of SPAICE2024: The First Joint European Space Agency / IAA Conference on AI in and for Space}, edited by D.~{Dold}, A.~{Hadjiivanov}, and D.~{Izzo}, 2024, pp. 247--251.
\newblock \doi{10.5281/zenodo.13885579}.

\bibitem[{Zucchelli et~al.(2024)Zucchelli, Wu, Briden, Hofmann, Rodriguez-Fernandez, and Linares}]{zucchelienrico2024finetunedlanguagemodelsasspacesystemscontrollers}
Zucchelli, E.~M., Wu, D., Briden, J., Hofmann, C., Rodriguez-Fernandez, V., and Linares, R., \enquote{Fine-Tuned Language Models as Space Systems Controllers,} \emph{Proceedings of the AAS/AIAA Astrodynamics Specialist Conference}, Broomfield, CO, 2024.

\bibitem[{Foutter et~al.(2024)Foutter, Bhoj, Sinha, Elhafsi, Banerjee, Agia, Kruger, Guffanti, Gammelli, D'Amico, and Pavone}]{foutter2024adaptingfoundationmodelspacebased}
Foutter, M., Bhoj, P., Sinha, R., Elhafsi, A., Banerjee, S., Agia, C., Kruger, J., Guffanti, T., Gammelli, D., D'Amico, S., and Pavone, M., \enquote{Adapting a Foundation Model for Space-based Tasks,} , 2024.
\newblock \urlprefix\url{https://arxiv.org/abs/2408.05924}.

\bibitem[{Wei et~al.(2023)Wei, Wang, Schuurmans, Bosma, Ichter, Xia, Chi, Le, and Zhou}]{wei2023chainofthoughtpromptingelicitsreasoning}
Wei, J., Wang, X., Schuurmans, D., Bosma, M., Ichter, B., Xia, F., Chi, E., Le, Q., and Zhou, D., \enquote{Chain-of-Thought Prompting Elicits Reasoning in Large Language Models,} , 2023.
\newblock \urlprefix\url{https://arxiv.org/abs/2201.11903}.

\bibitem[{Yao et~al.(2023)Yao, Zhao, Yu, Du, Shafran, Narasimhan, and Cao}]{yao2023reactsynergizingreasoningacting}
Yao, S., Zhao, J., Yu, D., Du, N., Shafran, I., Narasimhan, K., and Cao, Y., \enquote{ReAct: Synergizing Reasoning and Acting in Language Models,} , 2023.
\newblock \urlprefix\url{https://arxiv.org/abs/2210.03629}.

\bibitem[{Radford et~al.(2021)Radford, Kim, Hallacy, Ramesh, Goh, Agarwal, Sastry, Askell, Mishkin, Clark, Krueger, and Sutskever}]{radford2021learningtransferablevisualmodels}
Radford, A., Kim, J.~W., Hallacy, C., Ramesh, A., Goh, G., Agarwal, S., Sastry, G., Askell, A., Mishkin, P., Clark, J., Krueger, G., and Sutskever, I., \enquote{Learning Transferable Visual Models From Natural Language Supervision,} , 2021.
\newblock \urlprefix\url{https://arxiv.org/abs/2103.00020}.

\bibitem[{Alayrac et~al.(2022)Alayrac, Donahue, Luc, Miech, Barr, Hasson, Lenc, Mensch, Millican, Reynolds, Ring, Rutherford, Cabi, Han, Gong, Samangooei, Monteiro, Menick, Borgeaud, Brock, Nematzadeh, Sharifzadeh, Binkowski, Barreira, Vinyals, Zisserman, and Simonyan}]{alayrac2022flamingovisuallanguagemodel}
Alayrac, J.-B., Donahue, J., Luc, P., Miech, A., Barr, I., Hasson, Y., Lenc, K., Mensch, A., Millican, K., Reynolds, M., Ring, R., Rutherford, E., Cabi, S., Han, T., Gong, Z., Samangooei, S., Monteiro, M., Menick, J., Borgeaud, S., Brock, A., Nematzadeh, A., Sharifzadeh, S., Binkowski, M., Barreira, R., Vinyals, O., Zisserman, A., and Simonyan, K., \enquote{Flamingo: a Visual Language Model for Few-Shot Learning,} , 2022.
\newblock \urlprefix\url{https://arxiv.org/abs/2204.14198}.

\bibitem[{Li et~al.(2022)Li, Li, Xiong, and Hoi}]{li2022blipbootstrappinglanguageimagepretraining}
Li, J., Li, D., Xiong, C., and Hoi, S., \enquote{BLIP: Bootstrapping Language-Image Pre-training for Unified Vision-Language Understanding and Generation,} , 2022.
\newblock \urlprefix\url{https://arxiv.org/abs/2201.12086}.

\bibitem[{Liu et~al.(2023)Liu, Li, Wu, and Lee}]{liu2023visualinstructiontuning}
Liu, H., Li, C., Wu, Q., and Lee, Y.~J., \enquote{Visual Instruction Tuning,} , 2023.
\newblock \urlprefix\url{https://arxiv.org/abs/2304.08485}.

\bibitem[{Wang et~al.(2024)Wang, Shen, and Stadie}]{wang2024solving}
Wang, Z., Shen, R., and Stadie, B.~C., \enquote{Solving Robotics Problems in Zero-Shot with Vision-Language Models,} , 2024.
\newblock \urlprefix\url{https://openreview.net/forum?id=RQDuFF1rOn}.

\bibitem[{Allen et~al.(2023)Allen, Rachlin, Ruprecht, Loughran, Varey, and Viggh}]{allen2023spacegym}
Allen, R.~E., Rachlin, Y., Ruprecht, J., Loughran, S., Varey, J., and Viggh, H., \enquote{SpaceGym: Discrete and Differential Games in Non-Cooperative Space Operations,} \emph{2023 IEEE Aerospace Conference}, IEEE, 2023, pp. 1--12.

\bibitem[{Akhtar et~al.(2023)Akhtar, Shankarampeta, Gupta, Patil, Cocarascu, and Simperl}]{akhtar2023exploringnumericalreasoningcapabilities}
Akhtar, M., Shankarampeta, A., Gupta, V., Patil, A., Cocarascu, O., and Simperl, E., \enquote{Exploring the Numerical Reasoning Capabilities of Language Models: A Comprehensive Analysis on Tabular Data,} , 2023.
\newblock \urlprefix\url{https://arxiv.org/abs/2311.02216}.

\bibitem[{OpenAI(2024{\natexlab{b}})}]{openai2024o1}
OpenAI, \enquote{O1: High-Precision Numerical Reasoning Model,} , September 2024{\natexlab{b}}.
\newblock \urlprefix\url{https://openai.com/o1/}, accessed: 2024-11-29.

\bibitem[{Savery et~al.(2021)Savery, Rogel, and Weinberg}]{xArm7inproceedings}
Savery, R., Rogel, A., and Weinberg, G., \enquote{Emotion Musical Prosody for Robotic Groups and Entitativity,} 2021, pp. 440--446.
\newblock \doi{10.1109/RO-MAN50785.2021.9515314}.

\bibitem[{Keselman et~al.(2017)Keselman, Woodfill, Grunnet-Jepsen, and Bhowmik}]{keselman2017intelrealsensestereoscopicdepth}
Keselman, L., Woodfill, J.~I., Grunnet-Jepsen, A., and Bhowmik, A., \enquote{Intel RealSense Stereoscopic Depth Cameras,} , 2017.
\newblock \urlprefix\url{https://arxiv.org/abs/1705.05548}.

\bibitem[{Dao et~al.(2022)Dao, Fu, Ermon, Rudra, and Ré}]{dao2022flashattentionfastmemoryefficientexact}
Dao, T., Fu, D.~Y., Ermon, S., Rudra, A., and Ré, C., \enquote{FlashAttention: Fast and Memory-Efficient Exact Attention with IO-Awareness,} , 2022.
\newblock \urlprefix\url{https://arxiv.org/abs/2205.14135}.

\end{thebibliography}

\end{document}